\documentclass[10pt,twocolumn,letterpaper]{article}

\usepackage{cvpr}
\usepackage{times}
\usepackage{epsfig}
\usepackage{graphicx}
\usepackage{amsmath}
\usepackage{amssymb}

\usepackage[breaklinks=true,bookmarks=false]{hyperref}

\cvprfinalcopy % *** Uncomment this line for the final submission

\def\cvprPaperID{****} % *** Enter the CVPR Paper ID here

\begin{document}

%%%%%%%%% TITLE
\title{Numerical Sequence Prediction using Bayesian Concept Learning\thanks{\rm As part of Computational cognitive modeling course project at NYU Courant}}

\author{Damarapati, Mohith ({\tt\small md4289@nyu.edu}),\\
Enaganti, Inavamsi Bhaskar ({\tt\small ibe214@nyu.edu}) and\\
Rajakumar, Alfred Ajay Aureate ({\tt\small aar653@nyu.edu})\\
Courant Institute of Mathematical Sciences, New York University, New York\\
% For a paper whose authors are all at the same institution,
% omit the following lines up until the closing ``}''.
% Additional authors and addresses can be added with ``\and'',
% just like the second author.
% To save space, use either the email address or home page, not both
%\and
}

\maketitle
%\thispagestyle{empty}

%%%%%%%%% ABSTRACT
\begin{abstract}
   When people learn mathematical patterns or sequences, they are able to identify the concepts (or rules) underlying those patterns. Having learned the underlying concepts, humans are also able to generalize those concepts to other numbers, so far as to even identify previously unseen combinations of those rules. Current state-of-the art RNN architectures like LSTMs perform well in predicting successive elements of sequential data, but require vast amounts of training examples.  Even with extensive data, these models struggle to generalize concepts.  From our behavioral study, we also found that humans are able to disregard noise and identify the underlying rules generating the corrupted sequences.  We therefore propose a Bayesian model that captures these human-like learning capabilities to predict next number in a given sequence, better than traditional LSTMs.
\end{abstract}

%%%%%%%%% BODY TEXT
\section{Introduction}

A sequence is a regularity with its elements repeating in a predictable manner.  Any sequence is built from a certain set of primitives which repeat according to a certain set of rules.  Humans are really good at observing these primitives and rules behind the patterns. Given a sequence, people could naturally predict what comes next by learning rules behind it. Current state-of-the-art Recurrent Neural Network architectures like Long Short-Term Memory (LSTM) networks require hundreds or thousands of data to predict the next number in a sequence as they fail to learn richer concepts which humans do very easily \cite{lstm}.

We would be dealing with strictly increasing mathematical sequences in this project.  Numbers in a mathematical sequence follow certain rules.  In general, a number in a sequence would depend on the number preceding it or its position. (Even if it's dependent on just the position in the sequence, the current number could still be related to its preceding number in some intuitive fashion).  But, in cases like Fibonacci series, a number depends on two numbers preceding it.  We propose a model that captures human learning abilities for predicting the next number in the sequence using Bayesian Concept Learning. 

In general, humans are remarkably adaptable to noisy sequences. People are fairly capable of identifying the underlying rules generating a sequence even when the sequence is corrupted with either stationary noise (where some elements are slightly away from their original hypothesis) or progressive noise (where the errors are propagating across a sequence). Our model also captures this ability of humans to perform well in a noisy environment, meaning identifying the underlying rule and thereby predicting the next number. In this report, we've considered only progressive noise in our Bayesian model while computing likelihood, as stationary noise might be harder to implement using our method. We claim that this type of Bayesian learning approach is similar to humans approach to capturing the rules of the intended sequence.

Over time, people gradually develop their sequence solving skills.  At first, they learn to predict sequences which just vary by a constant addition or multiplication factor.  As they develop, they could predict sequences with combinations/compositions of multiplication and addition factors.  And, when their expertise improves more, they could even predict sequences which depend on two preceding numbers.  We are able to capture this gradual learning process of humans in our model with the help of a parameter called \textbf{Human Experience Factor - $\beta_{hef}$}.

The core ideas and inspiration were from the paper on number game (Ref. \cite{tenen}) and (Ref. \cite{tenen2}). Although Ref. \cite{jaini} is somewhat remotely related, our work is pretty much outlined by Ref. \cite{Hutter}. As described in Ref. \cite{tenen2}, our models could also be formalised using $lambda$ functions, but we're not using that formalism here.

%------------------------------------------------------------------------
\section{Mathematical Sequence}

In a mathematical sequence, usually a number $x(n)$ depends either on just $n$ (or the position) or the previous element(s) in the sequence, $x(n-1),x(n-2),...$.  Here, for simplicity we are going to consider only sequences where each number depends on just the previous number.

Hence, the rule that we consider for generating sequences is:
$x(n)=f (x (n-1))$, where $f:\mathbb{N}\rightarrow\mathbb{N}$

Here we'll be considering only linear functions as generators or $f$. So,
$x(n)=ax(n-1)+b$, for some positive integers  $a$ and $b$.

Examples of sequences:

Ex. 1.1.  Rule: $x(n) = 2x (n-1)$;
Example of a sequence generated using this rule: $2, 4, 8, 16,...$

Ex. 1.2.  Rule: $x(n) = 2x (n-1) + 2$;
Example of a sequence generated using this rule: $4, 10, 22, 46,...$

Ex. 1.3.  Rule: $x(n) = x(n-1) + 5$;
Example of a sequence generated using this rule: $2, 7, 12, 17,...$

Here, you could notice that Ex 1.1. is a sequence of powers of two.

\section{Bayesian Concept Learning: Model Setup}
  
\subsection{Primitives and Hypothesis}

A lot of strictly increasing mathematical sequences can be generated by a just a series of multiplications ($\times$) and additions ($+$) of certain numbers. 

For the sake of simplicity, let us assume just the following 12 primitives, $10$ for addition ($+$) and $2$ for multiplication ($\times$). Let $A_{i}(p) = p+i,	i \in {1,2...10}$ (meaning add $i$ to the previous number)	
Similarly, let $M_{i}(p)=p  \times i,   i \in {2,3}$ 	 (meaning multiply $2$ or $3$ to the previous number)

For each of the above defined primitives, we will store a possible transition pair.  We assume these pairs as something that humans could infer right way from their elementary math knowledge or memory.

Examples for primitive pair lists:

Ex 2.1. $L(A_{1})  = \{ (1,2) , (2,3) ,..., (99,100) \}$

Ex 2.2. $L(M_{2}) = \{ (1,2) , (2,4) ,..., (49, 98) \}$

Ex 2.3. $L(A_{5})  = \{ (1,6) , (2,7) ,...,(94, 99 ) \}$

We generate different hypotheses by combining these primitives - combining one of the $M_{s}$'s with one of the $A_{t}$'s to obtain: $x_{n}=A_{t}(M_{s}(x_{n-1}))=s\times x_{n-1}+t$

Here,  $M_{s}$ is from the set of multiplicative primitives $\{M_{2}, M_{3}\}$ and $A_{t}$ is from the set of additive primitives $\{A_{1}, A_{2},...., A_{10}\}$ So, number of possible such compound hypotheses for a sequence would be $10 \times 2 = 20$.

A compound hypothesis is a combination of primitives (one additive and one multiplicative hypothesis combined). Examples of sequences generated from such compound hypotheses:

Ex. 3.1. $H$ with primitives - $(M_{2}, A_{2}) \rightarrow x_{n}=2x_{n-1}+2$.  Sequence example: $2, 6, 14, 30,...$
Ex. 3.2. $H$ with primitives - $(M_{3}, A_{5}) \rightarrow x_{n}=3x_{n-1}+5$.  Sequence example: $5, 10, 15, 20,...$

\subsection{Noise}

Before we discuss about the Bayesian method of concept learning let us assume that the training and test sequences could be corrupted by some noise. Here,w.l.o.g. we add standard Gaussian noise $N(0,\sigma)$ (properly normalized so that only integer values are allowed) to the sequence, which could be introduced in the following two ways:

\textbf{Stationary Noise} - Consider a noise free sequence generated from a hypothesis, and now to each term we add noise but with obviously only integer numbers in the sequence.

Example $M_{2} : 2, 4, 8, 16, 32, 64$.,  with noise might become $2, 5, 8, 16, 32, 64$.

\textbf{Progressive Noise} - Here, every time we generate the next term we add noise. Then, we take the term with noise added and continue with the generation.

Example $M_{2} : 2, 4, 8, 16, 32, 64$.,  with noise will become $2, 5, 10, 20, 40, 80$

For the sake of simplicity, we are only considering Progressive Noise in our project.  All the models can be effortlessly extended to Stationary Noise too.

\subsection{Prior}
The framework described here is closely related to the ones described in Ref. \cite{tenen}. We are assuming that the prior probability would depend on the human experience, meaning, it would be a function of how much familiar the human is to the mathematical sequences that we present here. So, we have the probabilities defined as a function of $\beta_{hef}$ which varies from $0$ to $1$ depending on whether the human is completely inexperienced or a math novice! Here we assume that for any $\beta_{hef}$ or human experience level, the probability of a human being able to identify hypothesis with just one primitive would be the same independent of whether the primitive is $A_{i}$ or $P_{i}$. But, depending on $\beta_{hef}$ or the human experience level, the probability of a human being able to identify the compound hypothesis (hypothesis formed by both $A_{i}$ and $P_{i}$) corresponding to the given sequence increases for higher values of $\beta_{hef}$.

The probability of choosing one of the primitive hypotheses would be, $P(h_{p\text{ }or\text{ } a})=1/[(n_{p}+n_{a})+\beta_{hef}n_{p}n_{a}]$
Similarly, the probability of choosing one of the compound hypothesis would be
$P(h_{c})=\beta_{hef}/[(n_{p}+n_{a})+\beta_{hef}n_{p}n_{a}]$, where $\beta_{hef}$ varies from $0$ to $1$; $n_{p}$ is the total number of product primitives and $n_{a}$ is the number of additive primitives and so $n_{p}n_{a}$ therefore represents the total number of compound hypotheses that are formed by combining a $A_{i}$ and a $P_{i}$.

\subsection{Likelihood} 
Our aim here is to compute $P(D|h_{i})$, where $D = x_{1}, x_{2}, x_{3} , x_{4}, ... ,x_{n}$ is the given sequence. Here $P(D|h_{i})$ is the probability that we get $D$ while using the generating rule $h_{i}$.

$P(D|h_{i}) = P(x_{1}|h_{i})\prod_{j=2}^{n} P( x_{j}|x_{j-1}, h_{i} )$

Here, $h_{i}$ is a combination of primitives.  
$h_{i} = \{A_{p}, M_{q}\}$, and

$P(x_{n}| x_{n-1},h_{i})   =  g(x_{n}-f(h_{i}(x_{n-1})))$, 
where $g(x)$ is a standard normal distribution $N(0,1)$ and $f(h_{i}(x(n-1)))=A_{p}(M_{q}(x(n-1)))$.\\
$M_{q}(k)$ is transition of number $k$ in $L(M_{q})$.\\
$A_{p}(k)$ is transition of number $k$ in $L(A_{p})$

(We assume here w.l.o.g., that $P(x_{j}|h_{i})=k$, some constant. We'll see that this constant would get cancelled out in most of our calculations, especially due to Bayesian, and so, it need not be included in our analysis.)

\subsection{Posterior}

Using Bayes’ Theorem :

$P(h_{i}|D) = P(D|h_{i})P(h_{i})/P(D)$

To find the most likely hypothesis $h^{*} : 
h^{*} = \max_{i} P ( h_{i} | D )$

After finding $h^{*}$, we generate the next number by just applying the hypothesis on the last element of
$h^{*} = \{M_{p}, A_{q}\}$.\\
$x(n+1) = p x (n) +q$

\subsection{Human Experience Factor ($\beta_{hef}$)}

We introduce a new parameter to the model which is human experience factor ($\beta_{hef}$).  The $\beta_{hef}$ is higher for people who are better at identifying sequences, because of "familiarity".  To capture it in the model, we assume that the prior probabilities are a function of $\beta_{hef}$, as mentioned in an earlier section.  For lower $\beta_{hef}$ values, prior probabilities for hypothesis with $\{A_{1}, A_{2},...., A_{10}, M_{2}, M_{3}\}$ will be high and for compound hypothesis, it would be low. As $\beta_{hef}$ increase, probabilities for compound hypothesis also becomes comparable to the primitive hypotheses.

\section{Recurrent Neural Networks}

We performed experiments specifically on LSTMs.  LSTM is a variant of an RNN. They are good at maintaining long term dependencies and hence they are perform well as sequence models. A brief description of a LSTM cell and related equations are explained below.
LSTMs are first introduced by S. Hochreiter and J. Schmidhuber in 1991 (Ref.\cite{lstm}). Gates are the major components of an LSTM which are input gate - $i(t)$, forget gate - $f(t)$, output gate – $o(t)$,input modulation gate - $g(t)$ and $c(t)$ - memory cell. At each time step, an LSTM cell takes $x(t)$, $h(t-1)$ and $c(t-1)$ as inputs and outputs $h(t)$ and $c(t)$. Working of an LSTM cell is described using sequence of equations below.

\begin{equation}
\begin{split}
    i(t) &= \sigma (W_{xi}x(t) + W_{hi}h(t-1) + b_{i})\\
    f(t) &= \sigma (W_{xf}x(t) + W_{hf}h(t-1) + b_{f})\\
    o(t) &= \sigma (W_{xo}x(t) + W_{ho}h(t-1) + b_{o})\\
    g(t) &= \phi (W_{xc}x(t) + W_{hc}h(t-1) + b_{c})
\end{split}
\end{equation}
\begin{equation}
\begin{split}
    c(t) &= f(t)\cdot c(t-1) + i(t)\cdot g(t)\\
    h(t) &= o(t)\cdot \phi(c(t))
\end{split}
\end{equation}

Here $\sigma(x)=\frac{1}{1+e^{-x}}$ (sigmoid activation) and $\phi(x)=\tanh{x}$ ($\tanh$ activation).

Memory gate enables LSTM to learn complex long- term temporal dependencies. Additional depth can be added by stacking these on top of each other.

\section{Behavioral Experiments}

% UPDATE THIS AND EXPLAIN MORE ABOUT BEHAVIORAL EXPERIMENTS

\begin{figure}
    \centering
       \includegraphics[width=1.0\linewidth]{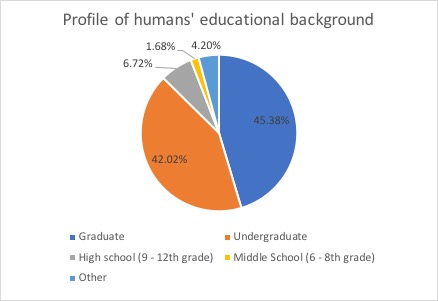}

    \caption{Pie-chart of the distribution of the participants' educational background}
    \label{fig:pie}
\end{figure}

We conducted an online survey, to observe how humans choose the next possible number among 4 options, more specifically, the $6^{th}$ number in a sequence where 5 numbers are given. About 119 people participated in the survey from different backgrounds. The profile of the participants is illustrated in Fig. [\ref{fig:pie}]. Since we passed it among our friends and family, we could observe that most of the participants are either graduates or undergraduates. Here, in our analysis, we are ignoring this selection bias and assuming that the participants would still be diverse enough in terms of identifying the underlying rules in a numerical sequence, as that depends more on the cognitive abilities of the humans which are very hard to profile.

The survey had 2 sections of 6 questions each. Each of those questions belong to a particular class/type of sequences. The sequences given in section 1 and their types are mentioned in table - 1. From about 54 participants (who opted to answer more questions), we received responses to 6 more questions of similar types (as the ones in section 1). So, overall we received about 173 responses to each of the 6 type of sequences.

\begin{table}
\begin{center}
\begin{tabular}{|c|l|}
\hline
Sequence Type & Sequence given in section 1\\
\hline\hline
& 11, 19, 27, 35, 43, \_\_\_ \\
Pure addition & a) 86; b) 51; c) 52; d) 53\\
& Ans: b) 51\\
\hline
& 5, 7, 9, 12, 14 \_\_\_ \\
Addition with noise & a) 17; b) 28; c) 12; d) 16\\
& Ans: d) 16\\
\hline
& 5, 10, 20, 40, 80 \_\_\_ \\
Pure multiplication & a) 120; b) 85; c) 160; d) 100\\
& Ans: c) 160\\
\hline
& 3, 6, 12, 25, 50, \_\_\_ \\
Multiplication & a) 75; b) 100; c) 56; d) 150\\
with noise & Ans: b) 100\\
\hline
 & 1, 4, 13, 40, 121, \_\_\_ \\
Pure compound & a) 122; b) 364; c) 243; d) 606\\
sequence & Ans: b) 364\\
\hline
& 5, 14, 33, 70, 144, \_\_\_ \\
Compound sequence & a) 580; b) 436; c) 148; d) 292\\
with noise & Ans: d) 292\\
\hline
\end{tabular}
\end{center}
\caption{Different class of sequences given to people in the survey. The sequences were given with and without noise or error as can be observed from the sequences in the 2nd column. The options that were given in the survey and the correct option are also given in the same column.}
\end{table}

\section{Experimental Results}

%-------------------------------------------------------------------------
\subsection{Results of the online survey}

\begin{figure}
    \centering
       \includegraphics[width=1.0\linewidth]{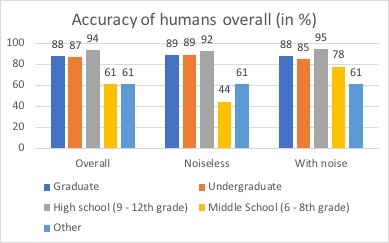}

    \caption{Overall accuracy of humans (left) from different education backgrounds, in predicting the next number in a sequence; the set of bars in the middle and in the right end show the accuracy of the participants in sequence prediction tasks when there is no noise in the sequence and when there is an error/noise in the sequence, respectively.}
    \label{fig:overall}
\end{figure}

From fig. \ref{fig:overall}, it is evident that humans are able to detect the generative rules of both noisy and noiseless sequences with similar precision. Humans are able to capture the concepts despite corruption of data with noise. As expected humans with more experience (or education qualification) were able to more easily understand the compound sequences.

\begin{figure}
    \centering
       \includegraphics[width=1.0\linewidth]{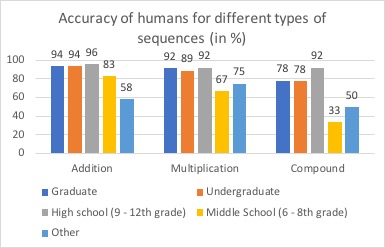}

    \caption{Accuracy of humans from different education backgrounds, while predicting the next number in a sequence. The set of bars to the left corresponds to sequences that are generated using just addition operation; similarly the set of bars in the middle corresponds to sequences that are generated using just multiplication operation and the set of bars in the right corresponds to sequences that are generated using both multiplication and addition operations.}
    \label{fig:types}
\end{figure}

%-------------------------------------------------------------------------
\subsection{LSTMs}

\begin{figure}
    \centering
       \includegraphics[width=1.0\linewidth]{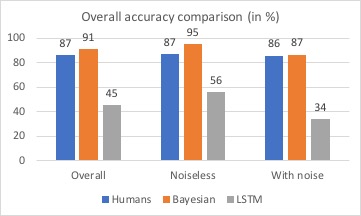}

    \caption{Comparing the overall accuracy (left) of humans with the overall accuracy of the Bayesian model and LSTM, in predicting the next number in a sequence. The set of bars in the middle and in the right end show the comparison of the accuracy of humans with that of the Bayesian model and LSTM for sequences that does not have errors and the noisy sequences (or the sequences with error), respectively.}
    \label{fig:comp_overall}
\end{figure}

\begin{figure}
    \centering
       \includegraphics[width=1.0\linewidth]{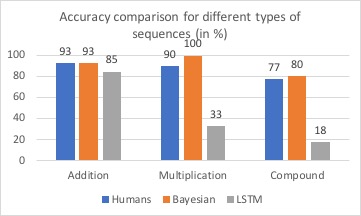}

    \caption{Comparing the accuracy of humans from different education backgrounds, with the accuracy of the Bayesian model and LSTM, while predicting the next number in a sequence. The set of bars to the left compares the accuracy of humans to that of the Bayesian model and LSTM for  sequences that are generated using just addition operation; similarly the set of bars in the middle corresponds to sequences that are generated using just multiplication operation and the set of bars in the right corresponds to sequences that are generated using both multiplication and addition operations.}
    \label{fig:comp_types}
\end{figure}

We trained LSTM models separately for additive, multiplicative and compound sequences with and without considering noise.  From our results, it is interesting to note that LSTMs are really good at learning addition without noise.  Our best LSTM model, with the hyper parameters as mentioned in table - 2, got an accuracy of $99.21\%$ without noise and $70.3\%$ with noise for addition (table - 3).  Although LSTMs performed well for addition, they  greatly under-performed when compared to humans for the multiplicative and compound hypotheses-based sequences (fig. \ref{fig:comp_types}). This brought down the overall accuracy of LSTMs and its accuracy for sequences with and without noise too (fig. \ref{fig:comp_overall}).

\textbf{Data} - Deciding right amount of data for a fair training of LSTM was a challenge for us.  We considered 999 as the maximum number for generating addition hypotheses to get a good amount of data to train our LSTM model.  But for the multiplication and combination hypotheses, 999 is a very low number and we could barely get sequences in that range.  Hence, we considered 99999 as our maximum number for multiplication and combination hypotheses. 
Another interesting aspect we considered for ensuring fair training of LSTM is to select a certain number of sequences such that probability of seeing all pairs is at least $70$.  We considered this to be a reasonable number for investigating LSTMs on our tasks.  More specifically, we chose $2500$ sequences out of $10000$ possible for training addition hypotheses, $500$ out of $1700$ possible for multiplication hypotheses and $300$ out of $1000$ possible for combination hypotheses. Despite giving the LSTM a slight advantage, their performance is still low for multiplication and combination hypotheses (fig. \ref{fig:overall} and table - 3).

\textbf{Data Encoding} - From our experiments, we observed that encoding of the data fed to LSTMs is an important factor while training LSTM.  We first followed one-hot encoding for each number in our dataset.  Results obtained were really bad and it was totally unfair for LSTMs to be trained that way.  We then considered one-hot encoding for each digit of a number which is introduced in Ref. \cite{graves}.  So, size of our encoding now became 30 for addition hypotheses (as maximum number is 999) and 50 for multiplication and combination hypotheses (as maximum number is 99999). 

\textbf{Hyper Parameters} - We considered a batch size of 16 with 2 hidden layers and trained all our models for 30 Epochs.  We followed 80-20 train test split (table - 2).

\begin{table}
\begin{center}
\begin{tabular}{|l|l|l|l|}
\hline
Hyper & Addition & Multipli- & Compound \\
parameters &&cation&\\
\hline\hline
Epochs & 30 & 30 & 30\\
\hline
Batch size & 16 & 16 & 16\\
\hline
No. of hidden & 2 & 2 & 2\\
layers &&&\\
\hline
Train data & 2500 & 500 & 300\\
\hline
MAX\_number & 999 & 99999 & 99999\\
\hline
\end{tabular}
\end{center}
\caption{Hyper parameters used in the LSTM implementation for different class of sequences.}
\end{table}

%-------------------------------------------------------------------------
\subsection{Bayesian Model}

Despite noise we noticed that the Bayesian model was able to very accurately predict the generating rule/hypothesis.

Here is an example. Consider the sequence : $1, 2, 3, 5, 8,...$
There are two highly probable rules that could have generated this sequence namely $(M_{2}, A_{1})$ with no noise or $(M_{2})$ with noise in each term.

Below are the results with the following parameters:
Noise mean = 0 \\
Noise variance at each element = 0.66 (This ensures ~half of the sequences are noiseless) \\
Prior of A, M, C : $(1,1,1)$ or $\beta_{hef}=1$ (meaning all the hypotheses including the compound hypotheses are equal) \\
\textbf{Result:} \\
\begin{equation}
\begin{split}
A_{2}  &:  0.74162549807248, \text{  Prediction :}  10 \\
A_{1}  &:  0.19419344602990238, \text{  Prediction :}  9 \\
M_{2}  &:  0.0639090892641122, \text{  Prediction :}  16 \\
A_{3}  &:  0.0002719666030357511, \text{  Prediction :}  11\\
\end{split}
\end{equation}

Here, 'Prediction' refers to the next possible number predicted by the model.

%-------------------------------------------------------------------------

\begin{table}
\begin{center}
\begin{tabular}{|l|l|l|l|}
\hline
Sequence Type & Humans & Bayesian & LSTM \\
&& Model & \\
\hline\hline
Pure addition & 94.22 & 99.99 & 99.21 \\
\hline
Addition with noise & 91.33 & 74.80 & 70.20\\
\hline
Pure multiplication & 91.33 & 99.99 & 48.00\\
\hline
Multiplication with & 88.44 & 99.99 & 18.00\\
noise &&&\\
\hline
Compound sequence & 76.88 & 56.42 & 21.00\\
\hline
Compound sequence & 78.03 & 89.11 & 15.00\\
with noise &&&\\
\hline
\end{tabular}
\end{center}
\caption{Comparing accuracy of humans (derived from the behavioral experiment) with that of the Bayesian model and LSTM for the 6 class/type of sequences. The accuracy mentioned here are in \%.
}
\end{table}

%-------------------------------------------------------------------------
\subsection{Impact of Human Experience Factor}

From the results, it can be seen that humans find it harder to solve compound sequences.  We noticed that experienced people were able to identify these compound sequences easily.  An expert tends to add more weight to the prior associated with the compound sequence.  We were able to capture this scenario using Human Experience Factor ($\beta_{hef}$).  In particular, we noticed that graduates were more comfortable with compound sequences compared to high school students in general.  This could possibly be a result of priors developing with experience.

\begin{figure}
    \centering
       \includegraphics[width=1.0\linewidth]{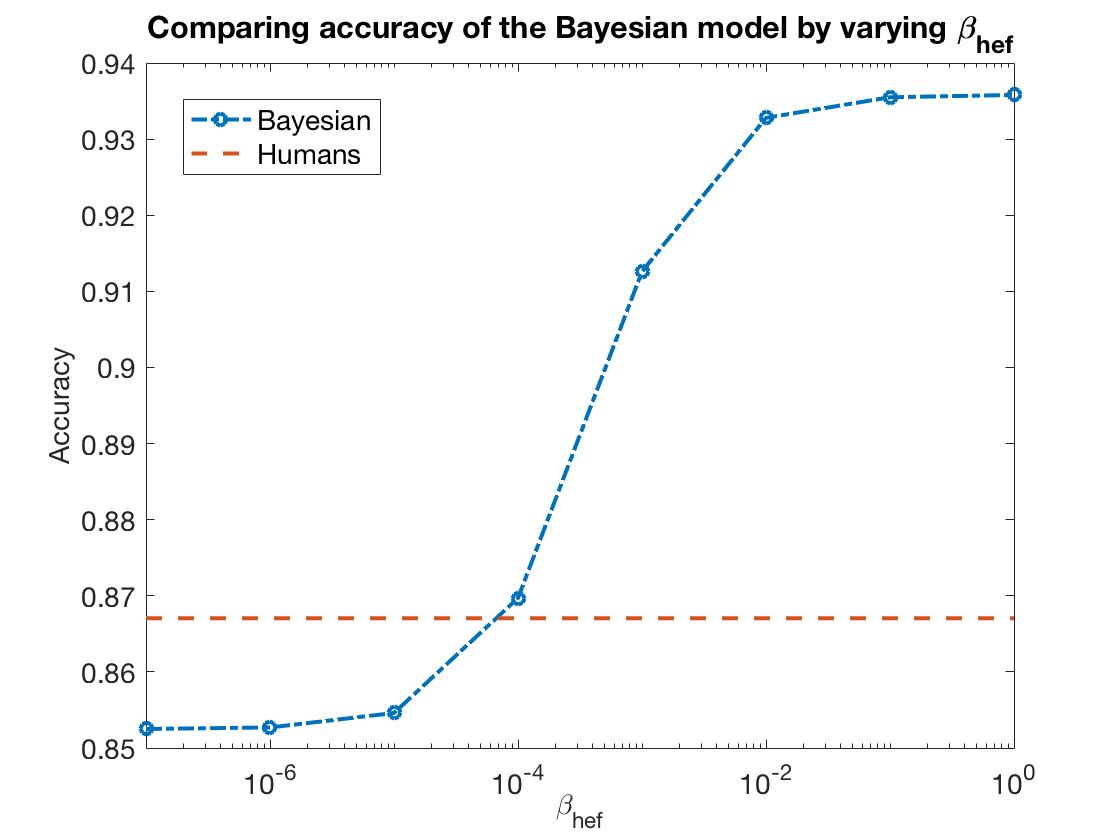}

    \caption{Comparing the accuracy of the Bayesian model with human accuracy for different values of $\beta_{hef}$. The x-axis is scaled logarithmically to illustrate the variation in accuracy while changing the $\beta_{hef}$ factor across orders of magnitude.}
    \label{fig:accu_comp}
\end{figure}

On another note, we tried to approximate the $\beta_{hef}$ factor averaging over a hundred humans as a group by comparing with our Bayesian learning model.  This plot is shown in Fig. [\ref{fig:accu_comp}]. The red line at $0.86705$ refers to the average human's accuracy, averaged over hundred participants. Whereas the blue line refers to the Bayesian model's accuracy for different values of  $\beta_{hef}^{eff}$. The point at which both the plots meet would correspond to the effective $\beta_{hef}^{eff}$ value for the entire group of participants as per the Bayesian model. It turns out be about $\beta_{hef}^{eff}=0.0001$. At that value of $\beta_{hef}=0.0001$, the overall accuracy of the Bayesian model matches with that of the group of human participants.

%------------------------------------------------------------------------
\section{Conclusion}

We noticed that LSTMs performed exceptionally well on noiseless addition and reasonably well on its noisy counterpart.  Despite LSTMs being able to grasp the additive rule, it failed to understand multiplication and compound sequences which are just meta-additive rules. 

As was expected, humans were able to figure out addition and multiplication sequences accurately and compound sequences to a reasonable extent.

The Bayesian learning model performed a lot better compared to LSTMs, specifically the multiplicative and compound sequences.  Despite the noise, the Bayesian model was able to accurately predict the generative rule for a sequence.  

From our behavioral data, we were able to gauge certain trends of human experience with sequences.  An higher Human Experience Factor ($\beta_{hef}$) implied increased familiarity with compound sequences.  From this we conclude that as humans come across more sequences, their priors also develop in such a way that they would be able to solve compound sequences.  This is analogous to how children improve their priors as they learn more patterns.  From our analyses, we observed that this kind of sequential concept learning does not seem to be captured by LSTMs.

%------------------------------------------------------------------------
\section{Future Work}
We have to admit that our survey results were skewed due to lack of proportionate representation of different groups.  We plan to rectify this with proper control groups. More steps have to be taken to compensate the inherent selection bias in a free-for-all survey!  

Our Bayesian model can be improved by using additional psychological and cognitive factors that might be affecting the way humans identify patterns in general. These may be derived from human data. It could be factors like, giving more weight to the latter numbers in the sequence, accounting for familiarity with round numbers ($+1, +2, +5, +10$) and inherent grouping of numbers in the sequence that were not a result of our hypotheses. For example: the sequence $3, 5, 7,...$ could either be identified as a sequence of odd numbers or a sequence of prime numbers).

We could also extend our model to more complex hypotheses, with very little modification. For instance, we could extend our model to Fibonacci-like sequences which depend on the previous two numbers unlike our hypotheses which only depended on the previous number.  

\long\def\acks#1{\vskip 0.15in\noindent{\large\bf Acknowledgments}\vskip 0.10in
\noindent #1}

\acks{We would like to acknowledge Prof. Brenden Lake and Reuben Feinman from NYU (USA) and Prabhu Prakash Kagitha from NIT Surat (India) for the useful discussions and suggestions.}

%{\small
%\bibliographystyle{ieee}
%\bibliography{egbib}
%}

\end{document}